# Machine Learning for Everyone: Simplifying Healthcare Analytics with BigQuery ML

[1]M.A. Salari and [2]B. Rahmani

[1]Saint Louis University, Computer Science Department, Saint Louis, USA

[2]Saint Louis University, Health & Clinical Outcome Research, Saint Louis, USA

**Abstract:** Machine learning (ML) transforms healthcare by enabling predictive analytics, personalized treatments, and improved patient outcomes. However, traditional ML workflows often require specialized skills, infrastructure, and resources, limiting accessibility for many healthcare professionals. This paper explores how BigQuery ML Cloud service helps healthcare researchers and data analysts to build and deploy models using SQL, without need for advanced ML knowledge. Our results demonstrate that the Boosted Tree model achieved the highest performance among the three models making it highly effective for diabetes prediction. BigQuery ML directly integrates predictive analytics into their workflows to inform decision-making and support patient care. We reveal this capability through a case study on diabetes prediction using the Diabetes Health Indicators Dataset. Our study underscores BigQuery ML's role in democratizing machine learning, enabling faster, scalable, and efficient predictive analytics that can directly enhance healthcare decision-making processes. This study aims to bridge the gap between advanced machine learning and practical healthcare analytics by providing detailed insights into BigQuery ML's capabilities. By demonstrating its utility in a real-world case study, we highlight its potential to simplify complex workflows and expand access to predictive tools for a broader audience of healthcare professionals.

## 1. Introduction

Artificial intelligence (AI) and machine learning (ML) are shaping the future of healthcare by applying advanced data analysis, predictive modeling, and decision support systems. The application of AI in healthcare allows for the identification of complex patterns in patient data, improving diagnostic accuracy, treatment personalization, and operational efficiency [1]. Healthcare providers are increasingly leveraging predictive analytics to foresee health outcomes, enabling earlier interventions and more targeted care [2][26]. For instance, AI models have proven effective in identifying high-risk patients and optimizing preventive care strategies [3]. Diabetes, a major global health challenge, requires early detection and preventive care. Predictive models built using accessible tools like BigQuery ML can help healthcare professionals identify at-risk individuals efficiently.

Cloud computing serves as a critical tool for AI and ML in healthcare, addressing many of the technical and infrastructural challenges associated with large-scale data analysis. With scalable infrastructure, cloud platforms allow healthcare providers to process and store vast amounts of data, facilitating AI-driven insights without the need of extensive on-site resources [4]. Cloud computing reforms healthcare by reducing costs, enabling scalability, improving patient access through telemedicine, and enhancing data-driven decisions. It fosters collaboration with seamless data sharing, streamlines hospital operations, and boosts efficiency, care quality, and outcomes [5]. Additionally, cloud computing in healthcare offers scalability and AI integration with addressing critical security and privacy challenges. Protecting sensitive patient data requires implementing regulatory-compliant measures and robust models. Research highlights various approaches to managing evolving service models and ensuring secure and efficient healthcare systems [6].

Among major cloud providers, Amazon Web Services (AWS), Microsoft Azure, and Google Cloud Platform (GCP) offer tools for AI and ML, with high strengths. AWS provides a wide range of machine learning tools and services such as SageMaker, which enables rapid model development and deployment. Microsoft Azure's AI services, including Azure Machine Learning, are known for their integration with the broader Microsoft ecosystem, which can be advantageous for healthcare organizations used for Microsoft products [7]. Google Cloud Platform focuses on data and machine learning services and is a powerful data warehouse simplifying the machine learning process of users familiar with data analytics [8].

Despite advancements in AI and ML, the need for user-friendly tools that simplify complex workflows remains critical. Traditional machine learning methods often require extensive coding skills, which limits their accessibility to a broader audience of healthcare practitioners. Cloud-based ML platforms provide user-friendly tools that reduce the need for specialized expertise and technical resources [9]. Through this integration, healthcare organizations can develop, deploy, and scale ML models with minimal local infrastructure, enhancing their ability to derive information from data [10].

The novelty of this study lies in demonstrating the use of BigQuery ML to enable healthcare professionals without programming expertise to build, train, and evaluate machine learning models directly through SQL queries. Unlike existing cloud-based platforms, BigQuery ML simplifies the integration of predictive analytics into healthcare workflows by eliminating the need for complex programming or data movement. This approach has not been explored in depth in prior research, which typically focuses on more programming-intensive platforms such as AWS SageMaker or Python-based workflows. The objective of this study is to demonstrate how BigQuery ML simplifies machine learning workflows in healthcare by enabling users to leverage SQL instead of traditional programming-based approaches. This approach reduces technical barriers, allowing healthcare professionals with limited programming expertise to utilize advanced ml capabilities for decision-making and predictive analytics.

This paper is structured as follows: Section 2 provides background on cloud-based ML solutions and their role in healthcare analytics. Section 3 (Methodology) describes the use of BigQuery ML for diabetes prediction, followed by a discussion on data preparation and preprocessing techniques. Model creation, training, and hyperparameter tuning in BigQuery ML are then presented, along with an evaluation of model performance using standard machine learning metrics. Section 4 (Results and Discussion) highlights key findings, including the strengths, limitations, and potential applications of BigQuery ML in healthcare. Section 5 (Future Directions and Usability Evaluation) discusses possible extensions of this study, including real-time predictive analytics and broader applications in healthcare. Finally, Section 6 (Conclusion) summarizes the key contributions and outlines future research directions.

Through this structure, we aim to highlight how BigQuery ML democratizes machine learning in healthcare, making predictive analytics more accessible, scalable, and efficient for a broader range of professionals.

## 2. Background

The intersection of AI and healthcare has been extensively studied, with particular emphasis on predictive analytics for chronic disease management. Diabetes, a global health challenge affecting millions, has been a focal point for such research due to its significant health and economic impacts. Numerous studies have demonstrated the potential of ML models in identifying diabetes risk using health indicators such as age, BMI, and blood pressure [11][12][13]. However, traditional ML approaches, including logistic regression and support vector machines, often require significant computational resources and specialized expertise, limiting their application in resource-constrained clinical environments.

Studies indicate that cloud-based ML has streamlined healthcare analytics by facilitating efficient data processing and model management [14]. While AWS and Azure provide comprehensive tools for machine learning, they require significant programming expertise and do not natively integrate SQL-based workflows. BigQuery ML, on the other hand, allows users to execute machine learning tasks within a single environment using SQL, making it particularly accessible for healthcare professionals without technical expertise. Table 1 summarizes the key features, advantages, and limitations of BigQuery ML compared to other leading platforms such as AWS SageMaker and Azure ML, highlighting its unique suitability for healthcare analytics.

**Table 1.** Comparison of Cloud-Based Machine Learning Platforms: Key features, advantages, and limitations of BigQuery ML, AWS SageMaker, and Azure ML, with a focus on accessibility, integration, and suitability for healthcare analytics.

| Service | Ease of Use | Integration | Expertise | Workflow | SQL Support | Scalability | Real-time | Healthcare Use |
|---|---|---|---|---|---|---|---|---|
| **BigQuery ML** | High (SQL) | Google Cloud | Low (SQL) | Simple | Yes | High | Yes | Strong |
| **AWS SageMaker** | Moderate (Python) | AWS (S3, Lambda) | High (Python) | Complex | No | High | Yes | Widely used |
| **Azure ML** | Moderate (GUI/Python) | Azure (Synapse) | Moderate | Mixed | No | High | Yes | Less accessible |

Platforms like BigQuery ML allow users to build and deploy machine learning models directly through SQL, reducing the technical barriers associated with traditional ML pipelines. BigQuery ML's ability to integrate seamlessly with healthcare datasets offers a powerful tool for healthcare professionals without programming expertise, democratizing access to advanced analytics. Despite these advantages, there is limited literature on the practical application of BigQuery ML in healthcare settings, particularly for large-scale datasets [8]. This study leverages the Diabetes Health Indicators Dataset to demonstrate the capabilities of BigQuery ML in a healthcare context. By utilizing SQL-based ML workflows, we aim to provide a framework for healthcare professionals to implement predictive analytics with minimal technical overhead, thereby addressing a critical gap in existing research.

*2.1 Related Studies and Literature Gap*

Most existing research relies on programming-intensive platforms requiring Python, R, or specialized ML frameworks, which create barriers for non-technical users. Limited studies explore SQL-based machine learning tools like BigQuery ML, which provide an accessible alternative for healthcare professionals lacking programming expertise. While studies have evaluated ML performance in healthcare using conventional programming methods, the accessibility and efficiency of SQL-driven ML workflows remain underexplored.

This gap is particularly significant given that SQL is widely used in healthcare data analytics [27][28][29], making it a natural choice for enabling predictive modeling without additional programming knowledge. Prior studies have not focused on how healthcare professionals can leverage SQL for ML without requiring external tools, manual data movement, or complex ML frameworks. BigQuery ML addresses this gap by allowing professionals to leverage the power of machine learning directly through SQL commands, making it an ideal solution for real-world healthcare applications. SQL remains one of the most widely used programming languages worldwide, consistently ranking among the top three in 2024 [27]. This widespread familiarity makes SQL a natural choice for professionals in healthcare analytics, enabling them to leverage existing skills for advanced machine learning tasks without requiring extensive technical training.

While alternative tools like Weka and ChatGPT offer unique approaches to machine learning workflows, they fall short in key aspects that are critical for real-world healthcare applications. Weka, for example, is a GUI-driven tool suitable for standalone ML studies but lacks integration with cloud platforms, making it inefficient for handling large-scale datasets or real-time analysis. Additionally, Weka requires users to export and import data manually, increasing the likelihood of errors and inefficiencies. Similarly, while ChatGPT can generate ML code through descriptive prompts, it assumes the user has a basic understanding of programming to interpret, test, and refine the code. Furthermore, ChatGPT does not provide a cloud environment to execute the generated code or access integrated tools such as BigQuery, Cloud Storage, or Vertex AI. In contrast, BigQuery ML operates within the Google Cloud ecosystem, enabling direct interaction with databases and other cloud tools. This capability streamlines the entire data workflow, from ingestion to model training and deployment, without requiring external environments or tools.

This study addresses this gap by showcasing BigQuery ML as a fully integrated cloud-based ML solution, demonstrating how it can be applied to structured healthcare data without requiring programming expertise or ML-specific knowledge.

*2.2 Research Question*

Given the increasing demand for accessible ML solutions in healthcare and the gap in research on SQL-driven ML platforms, this study aims to address the following research question:

"How effectively can BigQuery ML be utilized for predictive healthcare analytics, and how does it compare in terms of accessibility, ease of use, and performance to other cloud-based ML solutions?"

This question is driven by the need to evaluate the feasibility of SQL-based ML workflows for healthcare professionals, as well as the potential advantages and limitations of using BigQuery ML for predictive analytics. By investigating this question, the study seeks to determine whether BigQuery ML can lower technical barriers to machine learning adoption in healthcare, providing a scalable, efficient, and accessible alternative to programming-heavy ML platforms.

## 3. Methodology

This section describes the end-to-end workflow for implementing machine learning models using BigQuery ML within Google Cloud Platform (GCP) for healthcare analytics. We first introduce the cloud infrastructure used for data storage, processing, and model execution, followed by an explanation of the BigQuery data warehouse and its role in handling large-scale structured healthcare datasets. Next, we outline the ML models employed—Logistic Regression, Boosted Tree, and Deep Neural Network (DNN)—and justify their selection for this study. Additionally, we provide details about the dataset, including its characteristics, preprocessing considerations, and rationale for use. The section concludes with a discussion on hyperparameter tuning and model evaluation.

Figure 1 illustrates the overall workflow, from data ingestion to model training and evaluation.

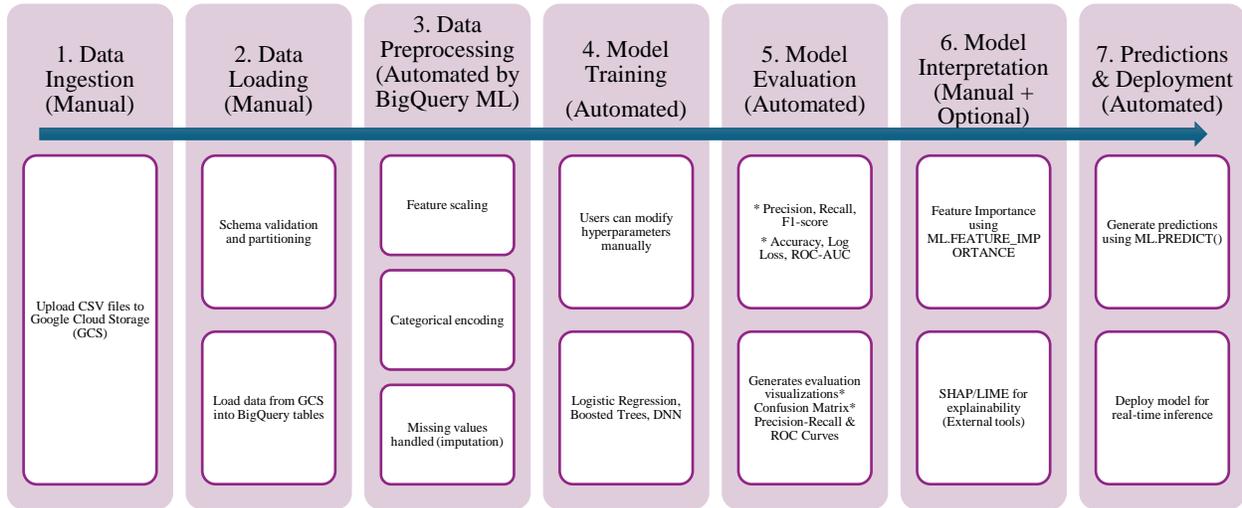

**Figure 1.** Workflow of BigQuery ML-based machine learning pipeline.

*3.1. Work Environment*

Google Cloud Platform (GCP) provides a robust, scalable cloud infrastructure for data storage, processing, and ML analysis. Its suite of tools enables end-to-end data workflows, from ingestion to predictive modeling, while ensuring compliance with healthcare regulations such as HIPAA [15]. The Google Cloud ecosystem, including BigQuery, Cloud Storage, and BigQuery ML, was chosen for this study due to its efficiency in handling large-scale healthcare datasets securely and seamlessly [16].

BigQuery, a fully managed serverless data warehouse, was utilized for data storage, querying, and preprocessing. Its low-latency SQL-based architecture is particularly well-suited for healthcare analytics, allowing rapid data exploration without the need for additional infrastructure management [17]. By storing the Diabetes Health Indicators Dataset in BigQuery, we leveraged its built-in preprocessing capabilities while maintaining data security and integrity.

BigQuery ML was used to build, train, and evaluate ML models directly through SQL queries. This SQL-based approach democratizes ML access, allowing healthcare professionals without programming expertise to implement predictive analytics with minimal overhead [18].

Hyperparameter tuning was conducted through iterative SQL-based adjustments in BigQuery ML. Learning rate, maximum iterations, and tree depth were optimized using ML.EVALUATE functions, ensuring a balance between precision and recall for healthcare predictions. Since BigQuery ML operates as a fully managed cloud service, no on-premise hardware was required for model execution. The computational workload was handled by Google Cloud's distributed infrastructure, leveraging high-performance computing resources optimized for SQL-based ML processing. This eliminates the need for dedicated GPUs or TPUs typically required for deep learning tasks, reducing hardware dependencies while ensuring scalability for large datasets.

*3.2. Dataset Used and Rationale*

The Behavioral Risk Factor Surveillance System (BRFSS) is an annual health survey conducted by the CDC since 1984. This telephone survey collects health-related data from over 400,000 Americans each year, focusing on health risk behaviors, chronic conditions, and preventive healthcare services. For this study, we used a refined 2015 BRFSS dataset, publicly available in CSV format on Kaggle. The dataset includes responses from 70,692 individuals, evenly balanced between non-diabetic (0) and pre-diabetic or diabetic (1) individuals. The target variable for this study is Diabetes_binary, which indicates whether a respondent has diabetes (1) or not (0).

To build our predictive models, we selected 21 key health indicators relevant to diabetes prediction. These features include demographic factors, lifestyle habits, and clinical indicators that have been commonly linked to diabetes risk in prior research [22]. Table 2 provides a summary of the 21 selected features, their type (numerical/categorical), key statistics, and brief descriptions:

**Table 2: Summary of Selected Features in the Dataset**

| Feature | Description | Type |
| --- | --- | --- |
| **HighBP** | High blood pressure (1 = Yes, 0 = No) | Categorical |
| **HighChol** | High cholesterol (1 = Yes, 0 = No) | Categorical |
| **CholCheck** | Cholesterol check in past 5 years (1 = Yes, 0 = No) | Categorical |
| **BMI** | Body Mass Index | Numerical |
| **Smoker** | Smoked at least 100 cigarettes in lifetime (1 = Yes, 0 = No) | Categorical |
| **Stroke** | History of stroke (1 = Yes, 0 = No) | Categorical |
| **HeartDiseaseorAttack** | Coronary heart disease or heart attack (1 = Yes, 0 = No) | Categorical |
| **PhysActivity** | Engages in physical activity outside work (1 = Yes, 0 = No) | Categorical |
| **Fruits** | Consumes fruit daily (1 = Yes, 0 = No) | Categorical |
| **Veggies** | Consumes vegetables daily (1 = Yes, 0 = No) | Categorical |
| **HvyAlcoholConsump** | Heavy alcohol consumption (1 = Yes, 0 = No) | Categorical |
| **AnyHealthcare** | Has healthcare coverage (1 = Yes, 0 = No) | Categorical |
| **NoDocbcCost** | Couldn't see doctor due to cost (1 = Yes, 0 = No) | Categorical |
| **GenHlth** | General health rating (1 = Excellent, 5 = Poor) | Ordinal |
| **MentHlth** | Days of poor mental health in past month (0–30) | Numerical |
| **PhysHlth** | Days of poor physical health in past month (0–30) | Numerical |
| **DiffWalk** | Difficulty walking or climbing stairs (1 = Yes, 0 = No) | Categorical |
| **Sex** | Gender (1 = Male, 0 = Female) | Categorical |
| **Age** | Age category (1 = 18-24, 13 = 80+) | Ordinal |
| **Education** | Education level (1 = No school, 6 = College graduate) | Ordinal |
| **Income** | Income category (1 = < $10,000, 8 = > $75,000) | Ordinal |
| **Target** | **Description** | **Type** |
| **Diabetes_binary** | Diabetes status (1 = Diabetic/Pre-diabetic, 0 = Non-diabetic) | Categorical |

*3.3 Machine Learning Models*

To apply BigQuery ML for healthcare analytics, three predictive models were selected: Logistic Regression, Boosted Trees, and Deep Neural Networks (DNN). While BigQuery ML supports a range of models, these models span a spectrum of complexity, from interpretable linear models to advanced ensemble methods and deep learning architectures. Their selection is based on prior research in healthcare predictive modeling, where they have shown effectiveness in disease prediction tasks.

- Logistic Regression – Chosen as a baseline model due to its interpretability and effectiveness for binary classification, making it particularly useful in clinical decision-making [19].
- Boosted Trees – Selected for its ability to capture complex relationships between features and improve classification accuracy through an iterative learning process [20].
- Deep Neural Networks (DNN) – Applied to explore the potential of deep learning in detecting high-dimensional patterns within structured healthcare datasets [21].

Justification for model selection is that while BigQuery ML supports various machine learning models, these three were deliberately chosen to showcase its capabilities across different levels of complexity; Logistic Regression serves as an interpretable, easy-to-use model for healthcare professionals with minimal ML expertise. Boosted Trees provide an intermediate-level approach, offering improved predictive performance while remaining interpretable. Deep Neural Networks (DNNs) demonstrate how BigQuery ML can handle more complex, computationally demanding models without requiring external tools or custom infrastructure.

This selection highlights how BigQuery ML enables users to scale from simple ML applications to advanced AI-driven healthcare solutions within a SQL-based environment. Future work can explore additional models, but this study focuses on demonstrating BigQuery ML's ease of use and versatility in healthcare analytics.

*3.4. Data Loading*

The Diabetes Health Indicators Dataset was first uploaded from Google Cloud Storage (GCS) to BigQuery. This process involved multiple steps to ensure the data was correctly ingested and prepared for machine learning tasks. The following outlines the key stages:

**Uploading the Dataset to Google Cloud Storage:** The raw dataset, *diabetes_binary_health_indicators_BRFSS2015.csv*, was stored as a CSV file in an existing or newly created GCS bucket. This setup facilitated easy access for data ingestion into BigQuery, ensuring data security and scalability.

**Creating the BigQuery Dataset:** A new dataset named **diabetes_analysis** was created in BigQuery to store the data. Configuration details such as Dataset ID and Data Location were specified according to project requirements. This step established a structured repository for managing the dataset within BigQuery.

**Loading Data from GCS to BigQuery:** The CSV data was loaded into BigQuery by creating a table within the diabetes analysis dataset. Configuration settings included:

Source: GCS path

(e.g., gs://your_bucket/diabetes_binary_5050split_health_indicators_BRFSS2015.csv).

File Format: CSV

Destination Table: diabetes data.

BigQuery ML automatically partitions the data into training and evaluation sets during model training using parameters such as data_split_method and data_split_eval_fraction. This eliminates the need for manual data splitting and ensures a standardized approach to model evaluation. Additionally, BigQuery ML validates the schema to ensure compatibility during ingestion and training, reducing the likelihood of errors caused by data mismatches.

*3.5. Data Preparation*

**Schema Validation and Compatibility:** BigQuery ML validates the schema during ingestion and training to ensure compatibility between the dataset and the machine learning model. This step reduces the likelihood of errors caused by data mismatches, streamlining the overall process and maintaining data integrity. By leveraging Google Cloud Storage and BigQuery, the data preparation and loading process became highly efficient and scalable, laying a strong foundation for subsequent model creation and training tasks.

**Automatic Preprocessing in BigQuery ML:** One of the advantages of BigQuery ML is its built-in automatic preprocessing, which eliminates the need for extensive manual data preparation. Automatic preprocessing consists of missing value imputation and feature transformations, ensuring consistency and efficiency in model training.

**Missing Value Imputation:** BigQuery ML automatically imputes missing values based on data type. For example, numerical features are replaced with the mean value, while categorical features are assigned to a special missing category. This prevents model failures due to missing data and ensures robust training.

**Feature Transformations**: Feature Transformations: BigQuery ML automatically standardizes numerical features (zero mean, unit variance) for logistic regression and DNNs, ensuring consistent feature scaling. Boosted trees and random forests do not require standardization, as they handle raw feature values natively.

**Categorical Feature Encoding:** BigQuery ML supports various encoding methods for categorical variables, including one-hot encoding, dummy encoding, label encoding, and target encoding. The choice of encoding depends on the model type, and details are available in the official documentation [30].

**Manual Data Preprocessing using SQL:** While BigQuery ML automates preprocessing, additional SQL-based feature engineering techniques can be applied for greater customization. Missing values can be handled with SQL queries by applying imputation techniques—numerical variables can be filled using the median, and categorical variables can be encoded using one-hot encoding. Additionally, feature selection can be conducted by calculating correlations between each feature and the target variable (diabetes_binary) using SQL-based statistical functions. These features streamline the data preparation process, allowing all preprocessing steps to be completed within the BigQuery environment without requiring external tools or programming. Figure 2.a demonstrates a sample SQL query for handling missing values, while Figure 2.b shows an example of categorical encoding. Figure 2.c illustrates how SQL can be used to calculate correlations for feature selection.

```sql
SELECT *,
       COALESCE(age, 50) AS age_imputed
FROM diabetes_data;
```

**Figure 2.a:** Handling missing values by applying median imputation using SQL example.

```sql
SELECT *,
       CASE WHEN gender = 'Male' THEN 1 ELSE 0 END AS gender_male,
       CASE WHEN gender = 'Female' THEN 1 ELSE 0 END AS gender_female
FROM diabetes_data;
```

**Figure 2.b:** SQL-based one-hot encoding of categorical variables example.

```sql
SELECT CORR(feature_value, diabetes_binary) AS correlation
FROM diabetes_data;
```

**Figure 2.c:** Calculating feature correlation with the target variable using SQL example.

*3.5. Model Creation and Training*

We employed BigQuery ML to train three models: Logistic Regression, Boosted Tree, and Deep Neural Network (DNN), leveraging SQL for all training workflows. BigQuery ML enables healthcare professionals to build machine learning models using SQL commands, simplifying the process for users without programming expertise. Each model was trained on the Diabetes Health Indicators Dataset with 21 feature variables.

We explain the SQL query for the Boosted Tree model in this section as an example. The SQL query used to create and train the Boosted Tree model is shown in Figure 3. This query demonstrates the simplicity and flexibility of BigQuery ML's SQL-based approach for model creation.

```sql
1   CREATE OR REPLACE MODEL `project_id.dataset_id.diabetes_model`
2   OPTIONS(
3     model_type = 'boosted_tree_classifier',
4     input_label_cols = ['Diabetes_binary'],
5     data_split_method = 'RANDOM',
6     data_split_eval_fraction = 0.2,
7     max_iterations = 150,
8     learn_rate = 0.05,
9     min_rel_progress = 0.00001,
10    l1_reg = 0.1,
11    l2_reg = 2.0
12  ) AS
13  SELECT
14    *
15  FROM
16    `project_id.dataset_id.diabetes_data`
17  WHERE
18    Diabetes_binary IS NOT NULL;
```

**Figure 3.** SQL query used for creating and training the Boosted Tree model in BigQuery ML. The query includes key hyperparameters such as learning rate, maximum iterations, and regularization terms to optimize model performance.

Here is a line-by-line explanation of the SQL query:

**CREATE OR REPLACE MODEL**: This command creates a new machine learning model or replaces an existing one with the same name. The model is stored in the specified project and dataset under the name diabetes_model.

**OPTIONS Clause**: This section specifies the configuration settings for the Boosted Tree model:

**model_type = 'boosted_tree_classifier'**: Indicates that the model is a Boosted Tree classifier, designed for binary classification tasks.

**input_label_cols = ['Diabetes_binary']**: Specifies that the target column for prediction is Diabetes_binary.

**data_split_method = 'RANDOM'**: Instructs BigQuery ML to randomly split the data into training and evaluation sets.

**data_split_eval_fraction = 0.2**: Allocates 20% of the data for evaluation and 80% for training.

**max_iterations = 150**: Sets the maximum number of iterations for the boosting process to converge.

**learn_rate = 0.05**: Defines the learning rate, which controls the contribution of each tree to the final prediction.

**min_rel_progress = 0.00001**: Specifies the minimum relative progress required between iterations to avoid early stopping.

**l1_reg = 0.1**: Applies L1 regularization to the model to prevent overfitting by penalizing large coefficients.

**l2_reg = 2.0**: Applies L2 regularization, adding an additional penalty for large coefficients to enhance generalization.

**AS SELECT Clause**:

- **SELECT \***: Selects all columns from the dataset for model training, ensuring that all relevant features and the target variable are included.

- **FROM 'project_id.dataset_id.diabetes_data'**: Specifies the source table containing the dataset.

- **WHERE Diabetes_binary IS NOT NULL**: Filters out rows with null values in the target column (Diabetes_binary), ensuring clean data for training.

The Boosted Tree model was designed to handle complex relationships in the data, leveraging BigQuery ML's efficient infrastructure. The use of hyperparameters such as max_iterations, learn_rate, and regularization settings (L1 and L2) ensures the model achieves a balance between accuracy and generalization, critical for healthcare applications like diabetes prediction. This SQL-based approach demonstrates the accessibility of advanced machine learning methods for healthcare professionals without the need for programming expertise. Hyperparameter tuning in BigQuery ML was conducted using iterative adjustments to model parameters through the OPTIONS clause in SQL queries. For example, for the Boosted Tree model, parameters such as learning rate, max iterations, and tree depth were specified and adjusted iteratively. Each model was evaluated using the ML.EVALUATE function to identify the optimal settings based on metrics such as log loss and F1 score. This approach ensured that the models achieved a balance between precision and recall, critical for healthcare predictions.

## 4. Results and Discussion

This section evaluates the performance of Logistic Regression, Boosted Tree, and Deep Neural Network (DNN) models using key metrics like precision, recall, F1 score, ROC AUC, accuracy, and log loss. Section 4.2 explains these metrics and their relevance in healthcare analytics. Section 4.3 examines confidence threshold trade-offs, while Section 4.4 analyzes precision-recall curves. Section 4.5 discusses ROC curves and model discrimination ability. Section 4.6 provides a comparative analysis of model performance, summarized in Table 1. Finally, Section 4.7 explores practical implications, guiding model selection for healthcare applications.

*4.1. Evaluation Metrics*

Model evaluation is a critical aspect of the machine learning process, especially in healthcare applications where predictive accuracy has real-world implications. The performance of the models, Logistic Regression, Boosted Tree, and Deep Neural Network (DNN), was assessed using multiple metrics, including precision, recall, F1 score, ROC AUC, accuracy, and log loss. BigQuery ML provides these metrics along with visual tools such as confidence threshold plots, precision-recall curves, and ROC curves. These tools are particularly useful in healthcare contexts for exploring model performance across various thresholds and understanding trade-offs between sensitivity and specificity, offering deeper insights into the strengths and limitations of each model.

In BigQuery ML, each trained model is automatically saved within the specified dataset. This built-in storage allows users to retrieve performance metrics directly from the model and streamlines the evaluation process. BigQuery ML's ML.PREDICT function also allows users to generate predictions on new or unseen data directly within BigQuery, integrating predictive analytics seamlessly into healthcare workflows.

The key evaluation metrics used in this study include:

1. **Precision (*P*)**: Measures the proportion of correctly predicted positive cases among all positive predictions. High precision reduces false positives [23].

$$Precision = \frac{TP}{TP + FP}$$

where *TP* is true positives and *FP* is false positive.

2. **Recall (Sensitivity, *R*)**: Represents the proportion of actual positive cases correctly identified by the model. High recall ensures minimal false negatives [23].

$$Recall = \frac{TP}{TP + FN}$$

where *FN* is false negatives.

3. **F1 Score**: The harmonic means of precision and recall, balancing the two metrics [24][25].

$$F1\ Score = 2 \times \frac{Precision \times Recall}{Precision + Recall}$$

4. **Accuracy (*A*)**: The ratio of correctly predicted cases (both positive and negative) to the total cases.

$$Accuracy = 2 \times \frac{TP + TN}{TP + TN + FP + FN}$$

where *TN* is true negative.

5. **Log Loss**: A metric used for probabilistic predictions, penalizing incorrect confidence levels. Lower log loss indicates better performance.

$$Log\ Loss = -\frac{1}{N} \sum_{i=1}^{N} ([y_i \log(\hat{y}_i) + (1 - y_i) \log(1 - \hat{y}_i)])$$

N represents the total number of samples, $y_i$ denotes the actual class label (0 or 1), and $\hat{y}_i$ refers to the predicted probability of the positive class (1).

6. **Receiver Operating Characteristic (ROC) AUC**: Measures the ability of the model to distinguish between classes across different thresholds. A higher AUC indicates better discrimination between positive and negative cases.

$$AUC = \int_{0}^{1} TPR(FPR) d(FPR)$$

These metrics provide a comprehensive evaluation of the models, ensuring a balanced assessment of their predictive performance in healthcare applications.

*4.2. Confidence Threshold vs Precision/Recall*

The leftmost panels in Figure 4 depict the relationship between confidence thresholds, precision, and recall for the three models. These graphs illustrate how the precision and recall values change as the confidence threshold is adjusted. Precision measures the proportion of true positive predictions among all positive predictions, while recall indicates the proportion of actual positive cases correctly identified by the model.

For the Boosted Tree model, precision and recall intersect at a balanced threshold, highlighting its ability to maintain a robust trade-off between the two metrics. The DNN model exhibits a similar pattern, with both precision and recall values demonstrating consistent performance as the threshold varies. In contrast, the Logistic Regression model shows a steeper decline in recall as the threshold increases, emphasizing its limitations in preserving sensitivity at higher confidence levels.

Understanding these trade-offs is crucial in healthcare, where high recall is often prioritized to minimize false negatives, such as undiagnosed diabetes cases. By analyzing these graphs, practitioners can identify thresholds that align with clinical priorities, ensuring that the selected model meets the specific needs of the application while balancing precision and recall effectively.

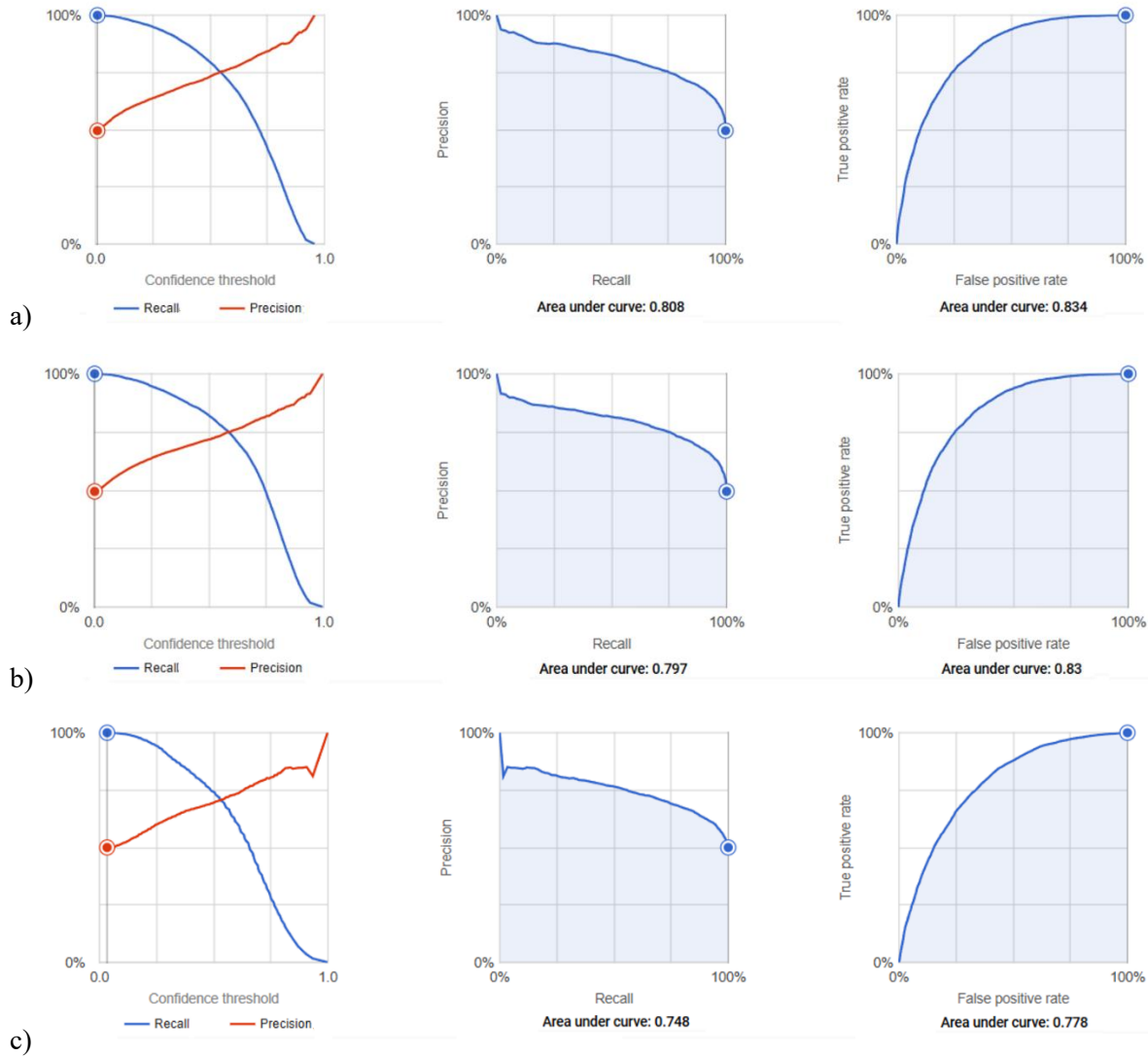

**Figure 4.** Model's precision-recall by threshold, precision-recall curve, and ROC curve. a) The Boosted Tree, b) DNN, c) The Logistic Regression.

*4.3. Precision-Recall Curve*

The middle panels in Figure 4 illustrate the precision-recall curves for the models. These curves provide a detailed view of the relationship between precision and recall across different thresholds. The area under the precision-recall curve (AUC-PR) is a critical metric for evaluating a model's ability to maintain high recall without sacrificing precision. For the Boosted Tree model, the AUC-PR is 0.808, indicating strong performance in identifying true positives while minimizing false positives. This robustness is particularly valuable in healthcare applications, where accurate identification of diabetes cases is essential. The DNN model achieved an AUC-PR of 0.797, slightly lower than the Boosted Tree but still demonstrates reliable performance. In contrast, the Logistic Regression model exhibited an AUC-PR of 0.748, highlighting its comparatively lower ability to balance precision and recall. Precision-recall curves are especially useful for imbalanced datasets, such as the diabetes dataset, where the positive class (diabetes cases) is relatively rare. By analyzing these curves, healthcare practitioners can gain insights into the models' trade-offs, enabling the selection of the most suitable model for their specific needs.

*4.4. Receiver Operating Characteristic (ROC) Curve*

The rightmost panels in Figure 4 show the Receiver Operating Characteristic (ROC) curves for the models. These curves are essential for evaluating a model's ability to distinguish between positive and negative classes by plotting the true positive rate (recall) against the false positive rate across various thresholds. The area under the ROC curve (AUC-ROC) provides a single, comprehensive metric to summarize the model's discriminatory power. For the Boosted Tree model, the AUC-ROC is 0.834, demonstrating its strong ability to differentiate between individuals with and without diabetes. This high value highlights the model's reliability in achieving a balance between sensitivity and specificity, critical for healthcare applications. The DNN model follows closely with an AUC-ROC of 0.83, showcasing comparable discriminatory power. The Logistic Regression model, while simpler, achieved an AUC-ROC of 0.778, indicating a modest but effective performance in distinguishing between the two classes. ROC curves are particularly valuable in healthcare scenarios where minimizing false negatives (e.g., missed diabetes cases) is essential. By analyzing the ROC curves, healthcare professionals can determine optimal thresholds that prioritize sensitivity while maintaining reasonable specificity, ensuring effective interventions for at-risk patients.

*4.5. Comparative Insights*

Among the evaluated models, the Boosted Tree model achieved the highest performance, with an F1-score of 0.7626 and an AUC-ROC of 0.8339 (see Table 1). These metrics were chosen to determine the best-performing model as they provide a balanced assessment of classification performance, particularly in healthcare applications where both false positives and false negatives must be minimized. The F1-score is widely recognized as an essential metric when dealing with class imbalance, as it accounts for both precision and recall. In diabetes prediction, missing a positive case (false negative) can lead to delayed diagnosis and worsened health outcomes, making recall particularly critical. However, an excessive focus on recall could lead to more false positives, which may overburden healthcare resources. Therefore, a model that optimally balances these factors—such as Boosted Tree—offers clinical and operational advantages in a real-world healthcare setting.

Similarly, AUC-ROC is a crucial metric for assessing a model's ability to distinguish between diabetic and non-diabetic patients across various decision thresholds. A higher AUC-ROC score indicates that the model consistently ranks positive cases above negative ones, making it particularly valuable for risk stratification in healthcare applications. The Boosted Tree model outperformed Logistic Regression and DNN in both F1-score and AUC-ROC, demonstrating better generalization and superior classification performance in this study. Based on these findings, the Boosted Tree model is recommended for diabetes prediction using BigQuery ML.

**Table 3.** Evaluation metrics of Logistic Regression, Boosted Tree, and DNN models trained in BigQuery ML.

|  | Precision | Recall | Accuracy | F1 Score | Log Loss | ROC AUC |
|---|---|---|---|---|---|---|
| Logistic Regression | 0.6972 | 0.7443 | 0.708 | 0.72 | 0.5674 | 0.7766 |
| Boosted Tree | 0.7324 | 0.7954 | 0.7546 | 0.7626 | 0.498 | 0.8339 |
| DNN | 0.7185 | 0.8215 | 0.752 | 0.7665 | 0.5049 | 0.8295 |

*4.6. Feature Importance Analysis for the Boosted Tree Model*

Feature importance helps us understand which variables contribute the most to the model's predictions. For the boosted tree classifier, we used importance gain as the primary metric. Importance gain measures how much a feature improves decision splits in the model, meaning higher values indicate more influential predictors. The results indicate that High Blood Pressure (HighBP) is the most significant predictor of diabetes, followed by General Health (GenHlth). This aligns with medical expectations, as high blood pressure and overall health perception are strongly associated with metabolic disorders, including diabetes. Other key features influencing diabetes prediction include Age, BMI, and High Cholesterol (HighChol), all of which are well-documented risk factors for diabetes. Difficulty Walking (DiffWalk), Heavy Alcohol Consumption, and History of Heart Disease or Attack also show moderate importance, reflecting broader cardiovascular risks associated with diabetes. Lifestyle factors such as Physical Activity, Smoking Status, and Vegetable Consumption have relatively lower importance in this model. While these factors influence overall health, their specific contribution to diabetes risk is smaller compared to the dominant predictors. To better illustrate the impact of each feature, we present a bar chart showing the importance gain of all features in Figure 5. Higher values indicate stronger predictive power.

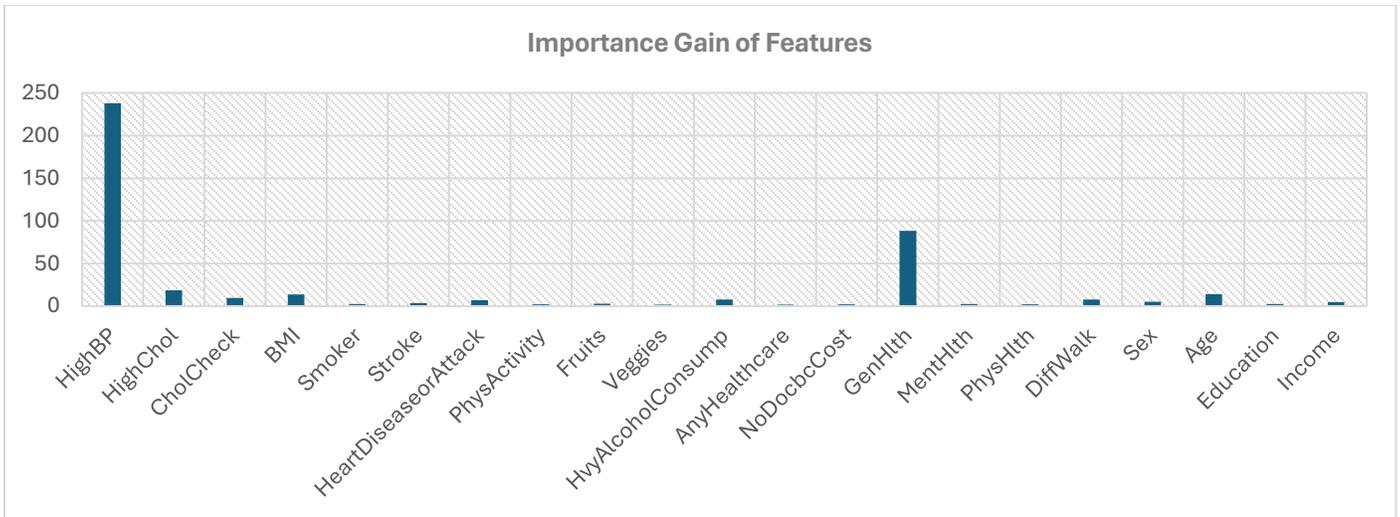

**Figure 5.** Feature Importance in Boosted Tree Model for Diabetes Prediction

In conclusion, the feature importance results align well with medical knowledge. Blood pressure, general health, age, BMI, and cholesterol levels play key roles in predicting diabetes. These insights can help refine predictive models and inform healthcare interventions.

*4.7. Discussion and Practical Implications*

*4.7.1 Strengths of BigQuery ML*

**Ease of Use:** The SQL-based interface of BigQuery ML allows users to perform end-to-end machine learning workflows—including data preparation, model training, evaluation, and prediction—entirely within BigQuery. This simplicity enables healthcare researchers, data analysts, and non-technical professionals to integrate predictive analytics into their decision-making processes without requiring advanced ML expertise. BigQuery ML also serves as a bridge for professionals transitioning into data engineering or data science roles by offering an accessible introduction to machine learning concepts using SQL. Non-technical managers and analysts can leverage BigQuery ML to generate insights from healthcare data without coding in Python or other programming languages.

**Scalability & Speed:** BigQuery ML's serverless architecture allows for the efficient processing of large-scale healthcare datasets in real time. Unlike traditional machine learning workflows that require manual ETL pipelines, BigQuery ML integrates data storage, preprocessing, and model training into a single environment, reducing the need for additional computational infrastructure. This scalability makes it suitable for population-level health studies, predictive modeling for disease risk assessment, and large-scale patient data analytics. By leveraging Google Cloud's infrastructure, healthcare organizations can process and analyze millions of records without significant latency. While BigQuery ML benefits from cloud scalability, organizations with cost or resource constraints may need to optimize queries and select cost-effective models to balance performance and affordability.

**Seamless Integration with Healthcare Workflows:** BigQuery ML's ability to integrate directly with Google Cloud services—such as Cloud Storage, Cloud Functions, and Vertex AI—enables a streamlined machine learning workflow for healthcare applications. This allows automated model updates, real-time predictions, and interoperability with electronic health records (EHRs) and other healthcare platforms. By reducing technical barriers and infrastructure overhead, BigQuery ML makes predictive healthcare analytics more accessible to hospitals, clinics, and public health organizations.

*4.7.2 Limitations and Challenges*

**Dataset Biases and Generalizability:** While this study utilizes a balanced diabetes dataset, healthcare datasets often suffer from biases related to demographics, socioeconomic factors, and healthcare accessibility. Since our dataset is derived from a specific population, it may not generalize well to other groups.

To mitigate bias and improve fairness, future research should explore external validation with diverse datasets to assess model robustness across different populations. Additionally, bias detection techniques and fairness-aware machine learning approaches could be integrated into BigQuery ML workflows.

**Interpretability of Complex Models:** While Boosted Trees and Logistic Regression models in BigQuery ML offer transparency and interpretability, Deep Neural Networks (DNNs) lack explainability, making them less suitable for high-stakes medical decision-making. To improve interpretability, techniques such as SHAP (Shapley Additive Explanations) and LIME (Local Interpretable Model-agnostic Explanations) can be integrated with BigQuery ML workflows. Future research can explore post-hoc interpretability frameworks that enhance model trustworthiness for healthcare applications.

**Image-Based Machine Learning Limitations:** BigQuery ML is optimized for structured, tabular datasets and does not natively support image-based machine learning. For healthcare applications involving medical imaging (e.g., X-rays, MRIs, CT scans), specialized tools such as Google Cloud AutoML Vision and TensorFlow should be considered. However, metadata associated with image datasets (e.g., diagnostic labels, patient history) can still be processed using BigQuery ML, providing valuable insights when combined with clinical data. Future work can explore hybrid workflows, where image analysis is performed using deep learning models in Vertex AI or TensorFlow, while BigQuery ML is used for structured data integration and analysis.

**Manual Hyperparameter Tuning:** BigQuery ML currently does not support automated hyperparameter tuning (e.g., grid search or random search). Instead, users must manually specify parameters in SQL queries, adjusting learning rates, iterations, and tree depths based on model performance. This manual tuning process may require additional effort compared to platforms like AWS SageMaker or Azure ML, which offer automated tuning features. Future versions of BigQuery ML could benefit from integrating AutoML-like hyperparameter optimization to further enhance usability.

## 5. Future Directions and Usability Evaluation

This study represents one of the first investigations into cloud-based SQL-driven machine learning workflows in healthcare. Future research should expand its applicability to a wider range of medical datasets and real-time predictive healthcare systems.

*5.1. Future Research Directions*

**Real-Time Predictive Analytics Integration:** Future work should explore the integration of BigQuery ML into real-time healthcare systems, such as hospital EHR systems, telemedicine platforms, and public health dashboards. Real-time analytics could enable immediate identification of high-risk patients, allowing timely interventions and improving patient outcomes.

**Expanding Dataset Diversity:** While this study focused on diabetes prediction, future research should explore broader healthcare challenges, such as cardiovascular diseases, cancer detection, and mental health.

Additionally, extending BigQuery ML's application to image-based medical datasets—such as X-rays, MRIs, and CT scans—could be an important area for further investigation. A hybrid approach integrating BigQuery ML for structured data with deep learning models in Vertex AI or TensorFlow for image analysis could enhance predictive insights and broaden the scope of healthcare analytics.

**Ethical Considerations and Bias Mitigation:** Ethical challenges, such as algorithmic bias and fairness, must be addressed in future research. Developing fairness-aware models and ensuring that BigQuery ML workflows are compliant with healthcare regulations (e.g., HIPAA) will be essential for equitable and responsible deployment.

*5.2. Possible Applications of the Research*

**Personalized Healthcare:** BigQuery ML can support the creation of personalized treatment plans by predicting individual risks for chronic conditions such as diabetes. High-risk patients could benefit from tailored interventions, including dietary adjustments, exercise regimens, and medication schedules.

**Population Health Management:** Public health organizations can leverage BigQuery ML to analyze population health trends, enabling resource allocation to communities with higher health risks. For example, targeted diabetes prevention campaigns could reduce disease prevalence and associated healthcare costs.

**Clinical Decision Support Systems (CDSS):** Integrating BigQuery ML predictions into CDSS could assist healthcare providers by flagging high-risk patients during routine checkups, reducing diagnostic delays and improving treatment outcomes.

**Remote Monitoring and Telemedicine:** BigQuery ML can analyze real-time data from wearable devices or remote monitoring systems, enabling continuous diabetes management and reducing hospital visits for chronic patients.

**Risk Assessment for Health Insurance:** Health insurers can use BigQuery ML to optimize risk assessments, personalize insurance plans, and incentivize preventive care, ultimately reducing costs and improving patient outcomes.

**Educational Tools for Healthcare Professionals:** BigQuery ML's SQL-based approach can serve as an educational tool for healthcare professionals, teaching them the fundamentals of predictive analytics and encouraging broader adoption of machine learning in clinical settings.

*5.3. Proposed Usability Evaluation*

While this study shows BigQuery ML's capabilities, a formal usability evaluation could provide deeper insights into its effectiveness for healthcare professionals. A structured questionnaire comparing BigQuery ML, WEKA, and Python-based tools could assess:

- Ease of Use: Comparing the SQL-based workflow with GUI-driven (WEKA) and code-based (Python) tools.
- Tool Preference: Evaluating user preference based on familiarity, flexibility, and usability.
- Cloud Services: Assessing the importance of scalability, integration, and real-time analytics in ML tool selection.
- Efficiency: Measuring workflow speed and reduction in external tool dependency.

- Overall Satisfaction: Rating usability and the likelihood of recommending BigQuery ML.

Such an evaluation would validate BigQuery ML's practicality in healthcare ML workflows and justify the benefits of SQL-based machine learning tools in cloud environments.

## 6. Conclusion

This study demonstrated how BigQuery ML simplifies machine learning workflows in healthcare by enabling predictive modeling through SQL-based queries. By eliminating the need for extensive programming expertise, BigQuery ML makes machine learning more accessible to healthcare professionals, data analysts, and researchers. Through a case study on diabetes prediction using the Diabetes Health Indicators Dataset, we evaluated three predictive models—Logistic Regression, Boosted Tree, and Deep Neural Networks (DNN)—and identified the Boosted Tree model as the best-performing approach based on F1-score and ROC AUC.

The results confirm that BigQuery ML's built-in preprocessing, scalability, and ease of integration with cloud-based healthcare data pipelines make it a valuable tool for predictive analytics. The feature importance analysis revealed that High Blood Pressure, General Health, BMI, and Age are significant predictors of diabetes, aligning with medical expectations. These findings underscore BigQuery ML's capability to support real-world healthcare applications, particularly in identifying high-risk individuals for early intervention. While BigQuery ML offers significant advantages in accessibility and efficiency, certain challenges remain, including model interpretability, dataset biases, and the lack of automated hyperparameter tuning. Future research should explore methods to enhance explainability, integrate real-time predictive analytics into healthcare systems, and expand the application of BigQuery ML to other diseases and medical datasets. The study highlights the broader impact of cloud-based machine learning on healthcare analytics, paving the way for more scalable, interpretable, and data-driven decision-making. By bridging the gap between advanced AI techniques and practical healthcare use cases, BigQuery ML provides a cost-effective and accessible solution for predictive analytics, supporting the future of personalized medicine and population health management.